# Two-Stage Hybrid Day-Ahead Solar Forecasting

Mohana Alanazi, Mohsen Mahoor, Amin Khodaei
Department of Electrical and Computer Engineering
University of Denver
Denver, USA
mohana.alanazi@du.edu, mohsen.mahoor@du.edu, amin.khodaei@du.edu

*Abstract*—Power supply from renewable resources is on a global rise where it is forecasted that renewable generation will surpass other types of generation in a foreseeable future. Increased generation from renewable resources, mainly solar and wind, exposes the power grid to more vulnerabilities, conceivably due to their variable generation, thus highlighting the importance of accurate forecasting methods. This paper proposes a two-stage day-ahead solar forecasting method that breaks down the forecasting into linear and nonlinear parts, determines subsequent forecasts, and accordingly, improves accuracy of the obtained results. To further reduce the error resulted from nonstationarity of the historical solar radiation data, a data processing approach, including pre-process and post-process levels, is integrated with the proposed method. Numerical simulations on three test days with different weather conditions exhibit the effectiveness of the proposed two-stage model.

*Keywords*—Solar generation forecast, global horizontal irradiance, autoregressive moving average model with exogenous input, nonlinear autoregressive neural network.

## Nomenclature

| | |
|---|---|
| $GHI_{actual}$ | Target global horizontal irradiance |
| $\overline{GHI}_{actual}$ | Average target GHI |
| $GHI_{forecast}$ | Forecasted global horizontal irradiance |
| $N$ | Number of GHI values |
| $q$ | Back shift operator |
| $t$ | Index for hour |
| $u(k)$ | ARMAX input |
| $v(k)$ | Disturbance/error |
| $y(k)$ | ARMAX output |
| $Y_t$ | Detrended solar time series |

## I. Intoduction

RENEWABLE generation has become a viable source that can provide sustainable and inexpensive supply of electricity, due to significant technological advances and many local and national incentives. However, generation from renewable resources has confronted variety of challenges, mainly because of the inherently variable generation. Such variability is caused by various climatic parameters such as temperature, air pressure, cloudiness, etc. [1]. An accurate forecast of generation of these resources will provide the power system operator the ability to plan ahead and control any generation variability form renewable resources by dispatching controllable generation resources in a coordinated fashion [1].

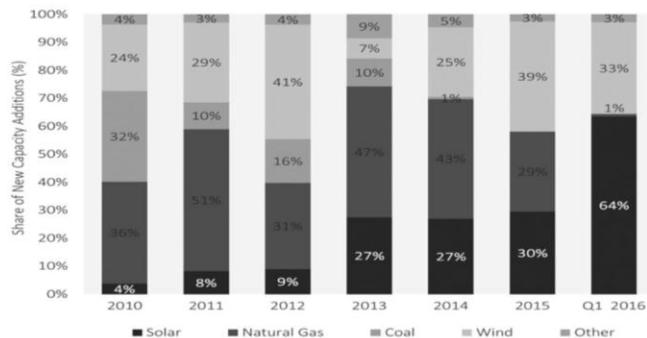

Fig. 1 The new added U.S. electric generation from 2010 to Q1 2016 [2].

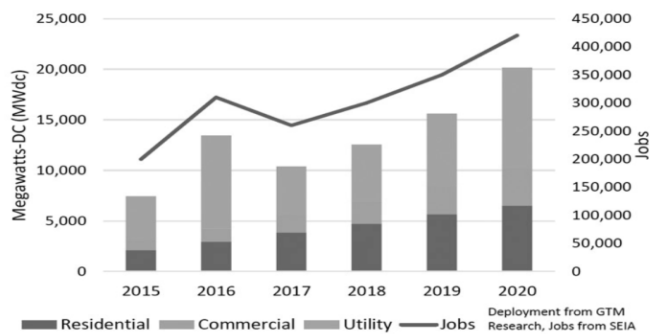

Fig. 2 U.S. solar energy deployment [3].

The growth in the amount of generation from renewable energy resources has been unprecedented. This growth is driven by the environmental concerns associated with $CO_2$ emissions and the global warming, as well as the state and governmental support of renewable resources, combined with the falling cost of the renewable energy technology. As shown in Fig. 1, in the first quarter of 2016 the added generation capacity from solar to the U.S grid represents 64% of the newly-added generation capacity. In the second quarter of 2016, the U.S. installed 2051 MW of solar, which is 43% higher of the installation in the similar timeframe in 2015, to reach a total installed capacity of 31.6 GW [2].

In December 2015, legislation was singed to extend the solar Investment Tax Credit (ITC) through 2020 in the Unites States. The ITC extension will result in more than 72 GW of PV to be deployed from 2016 through 2020. Fig. 2 depicts the projected solar PV deployment in this timeframe [3]. With such increase in solar generation installations, a proper forecasting is needed to help power system operators safely integrate solar generation and accordingly optimize electricity production and system management.

There are extensive studies on solar generation forecasting based on variety of methods with the common goal of minimizing the forecasting error. Using an efficient forecasting tool, power system operators will be able to schedule generation, obtain operating reserves, and administrate changes in loads and power outputs economically. Solar forecast is used in power industry to shape generation portfolios. For instant, a total of 38 GW of solar capacity was traded in energy market in Germany. Such amount of capacity would robustly have an effect on market prices [4]. Forecasting is commonly performed using physical or statistical models, or a combination of the two. Physical models rely on the physical description of the atmosphere and utilize the current observations of weather data and process them in order to predict future conditions using supercomputers. Physical models are suitable for long-time horizons and can be divided into two categories of numerical weather prediction (NWP) and the satellite/cloud imagery. The NWP is based on the current observations in atmosphere, which are processed to produce hundreds of meteorological elements such as temperature and solar irradiance through a process called assimilation. There are different NWP models such as global forecast system (GFS) and the ERA5 by the European center for medium range weather forecast [5]. In [6], different NWP models are analyzed to predict 14 hours of GHI, where the resulted root mean square error (RMSE) ranges from 20% to 40%. The satellite/cloud imagery helps understand the cloud motion by knowing the cloudiness with high spatial resolution. By understanding the cloud motion, the cloud position can be predicted, and thus the solar irradiance can be forecasted [7].

The statistical models require a large set of historical data in order to form a relationship between input and other important factors to forecast the output. These models rely on different mathematical algorithms to identify the patterns and trends in the time series. The common statistical models are persistence model, which predicts the next value based on the previous value, autoregressive (AR), moving average (MA), autoregressive and moving average (ARMA), and artificial neural networks (ANNs). Time series models aim to predict the future sequence of observations using historical data over various time horizons such as hourly, daily, or weekly. As the observations could be random, the time series is referred as a stochastic process [8]. Time series models focus only at the patterns of the data. In order to forecast a time series, these patterns should be identifiable and predictable. One of the most widely-used time series models is the ARMA model, which was created by Peter Whittle in 1951 and thoroughly developed and explained by Box-Jenkins in 1971. The ARMA model can be represented mathematically as $X_t = \sum_{i=1}^{p} \phi_i * X_{t-i} + \sum_{j=1}^{q} \beta_j * \varepsilon_{t-j} + \varepsilon_t$ , where $\phi_i$ is coefficient for AR part, $\beta_j$ is the coefficient for MA part, $\varepsilon_t$ is the white noise, and $p$ and $q$ are the orders of the AR and the MA, respectively. In [9], it is reported that the ARMA model shows an improvement in the mean square error (MSE) as much as 44.38% over the persistence model for 1-hour-ahead forecast. It should be noted that the time series has to be stationary before it is fed to the ARMA model [10]. The artificial neural network (ANN) is another viable statistical forecasting method that is based on the idea of the biological neural system in the human brain. The ANNs have the ability to process a complex nonlinear time series and find the relationship between the input and the target output using different training and learning algorithms. There are different types of ANNs such as the recurrent neural network (RNN), feed forward neural network (FFNN), and radial basis function neural network (RBFNN). A detailed review of different ANNs for solar forecasting applications is provided in [11]. Hybrid models have become more popular as they offer the combined advantages and reduce the limitations of other methods. In hybrid models, two or more forecasting methods are combined to get a better forecasting accuracy. In [12], a hybrid model of a variety of forecasting models is proposed to predict the next 48 hours solar generation in North Portugal. The hybrid model has shown an improvement of 57.4% over the persistence model and 34.06% over the statistical model. Some of the viable solar forecasting models in the literature can be found in [13]–[25].

The objective of this paper is to propose a two-stage model to improve the solar irradiance forecasting. In each stage, a certain method is used to provide a specific part of the forecast. As a contribution to this area of study, this paper develops the decomposed model in a way that nonlinear and linear models are separated, thus greatly improving individual as well as overall forecast. The accuracy of the two-stage model is addressed and compared with the single-stage method. To further improve the forecasts, a data processing approach is proposed that prepares and feeds stationary data to the two-stage forecasting model.

The rest of the paper is organized as follows: Section II presents the architecture and the formulation of the proposed forecasting model. Numerical simulations are presented in Section III. Discussions and conclusions are provided in Section IV.

II. THE ARCHITECTURE OF THE FORECASTING MODEL

Fig. 3 depicts the architecture of the proposed decomposed forecasting model. The forecasting model uses two cascaded stages based on Nonlinear Autoregressive Neural Network (NARNN) and the Autoregressive Moving Average with Exogenous Input (ARMAX). The main advantage of this decomposed model is to process both linear and nonlinear parts of the solar time series. NARNN deals with the nonlinear part of the forecasting and is used to predict a fitting model based on the historical stationary solar data. On the other hand, ARMAX considers the linear part of the forecasting and is used to forecast solar irradiance using the predicted fitting model as an input. These two stages along with data

processing is explained in the following:

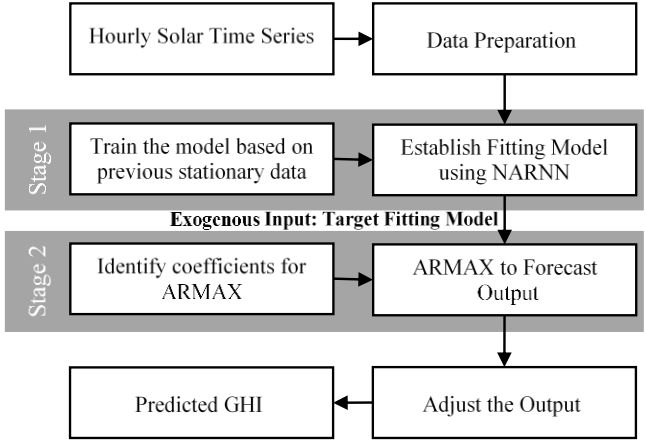

Fig. 3 The architecture of the proposed forecasting model.

### A. Data Prepration and Preprocessing

The solar data considered in this model are the horizontal global irradiance (GHI) that represent the solar irradiance received at horizontal surface on the ground. The historical GHI data are analyzed and undergone several preprocessing steps to ensure the quality of the data.

The GHI data preparation includes: removing GHI nighttime values, removing offset, and detrending. The offset is removed by subtracting the historical GHI from the clear sky GHI, which is the maximum GHI received at the surface in clear sky conditions. The resultant data from this process represent solar irradiance scattered by clouds or other factors. The resultant data are more dependent on the location and time that reflects other meteorological data. The next step is to consider only daytime hours, as solar output at nighttime hours is zero. By eliminating nighttime hours, knowing the exact sunrise and sunset times, the data size will be reduced to almost half, which accelerates time needed for simulation.

Statistical models require data set to be stationary before it is fed to forecasting tool. The output data from previous two steps are detrended and then tested using the Augmented Dickey–Fuller (ADF) test to validate the stationarity of the time series. The ADF test checks if there is a unit root. If test result comes below a defined critical value, then the null hypothesis should be rejected and the time series is stationary. Otherwise, the null hypothesis should be accepted and the time series is nonstationary. Different detrending models are addressed and compared in [26] and [27].

The data are detrended using Al-Sadah's model, which is represented mathematically in (1). Constants $a_0, a_1 \ldots a_n$ are determined using the least square regression analysis to fit the actual data set:

$$Y_t = a_0 + a_1 h + a_2 h^2 + \ldots + a_n h^n \qquad (1)$$

where $h$ is the local time. After the data are tested for the stationarity and verified, it will be normalized to obtain a number between 0 and 1. Normalization is an important step to ensure all data sets are under same reference scale, and to eliminate any variability due to the changes in the peak of the clear sky irradiance. More detail on GHI data preprocessing can be found in [28].

### B. Forecasting – Stage 1: Nonlinear Autoregressive Neural Network

The NARNN model is a time series model that requires a large set of historical data. In order to train the model and predict the fitting model, a large set of the previous hourly stationary data from the target day are used. One key issue is that the larger number of days that are used for the training, the more accurate the prediction will be. The NARNN is presented in (2) where $d$ is the number of previous hourly samples, determined through trial and error.

Once the fitting model is achieved, it is introduced as an exogenous input to the second stage of the forecasting, i.e. ARMAX. The predicted fitting model form this stage plays an active role in forecasting of the next stage, where a more accurately predicted fitting model in the first stage better ensures a more precise forecasting result in the second stage. The architecture of the NARNN is depicted in Fig. 4.

$$y(t) = f(y(t-1), y(t-2), \ldots, y(t-d)) \qquad (2)$$

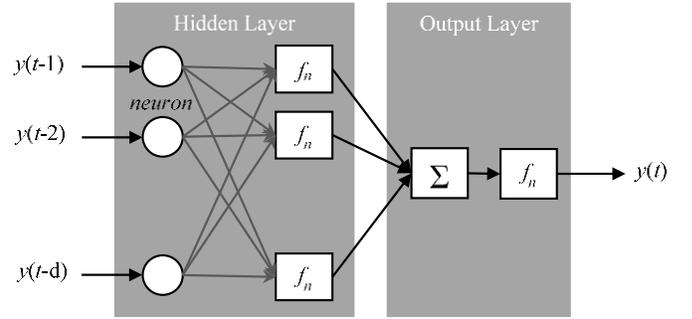

Fig. 4 The architecture of the NARNN.

In order to evaluate the performance of the fitting model the coefficient of determination $R^2$, which ranges between 0 and 1, is calculated as in (3). If $R^2 = 1$, the NARNN is able to predict the fitting model without any error, while $R^2 = 0$ means that the NARNN is not able to predict the fitting model and further training is needed.

$$R^2 = 1 - \left[ \frac{\sum_{t=1}^{N}(GHI(t)_{actual} - GHI(t)_{forecast})^2}{\sum_{t=1}^{N}(GHI(t)_{actual} - \overline{GHI(t)}_{actual})^2} \right] \qquad (3)$$

### C. Forecasting – Stage 2: Autoregressive Moving Average with Exogenous Input

The autoregressive moving average model with exogenous input includes two parts and can be mathematically represented as in (4).

$$A(q)y(k) = B(q)u(k) + C(q)v(k) \qquad (4)$$

where, $A(q)=1+a_1q^{-1}+\ldots+a_nq^{-n}$, $n$ is the order and $a_1,\ldots,a_n$ are coefficients for the AR part. $B(q)=b_1+b_2q^{-1}+\ldots+b_mq^{-m+1}$, $m$ is the order and $b_1,\ldots,b_m$ are coefficients for the input. $C(q)=1+c_1q^{-1}+\ldots+c_rq^{-r}$, $r$ is the order and $c_1,\ldots,c_r$ are coefficients for the MA part.

In order to find the coefficients for both AR and MA parts, the previous day is used to train the ARMAX model and estimate coefficients. The order of the ARMAX can be identified using the partial and autocorrelation plots. More detail on how to estimate the order of ARMAX model can be found in [29]. In addition, the Akaike Information Criterion (AIC) can be used to determine the order of ARMAX model. The AIC is modeled under different ARMAX orders and the best order is the one with the lowest AIC [30]. However, in this paper another method is used to find the orders of the AR and the MA. It is assumed that the orders of both the AR and the MA are the same, and then the error value is calculated by increasing the orders. The point with the least error for test data is considered as the best order for the ARMAX model. Fig. 5 shows the procedure of finding the order of the ARMAX model.

Once the ARMAX is developed, the fitting model predicted from Stage 1 is introduced as an exogenous input to this stage and the target output is forecasted. To achieve the final GHI, the forecasted output resulted from the ARMAX is adjusted as well.

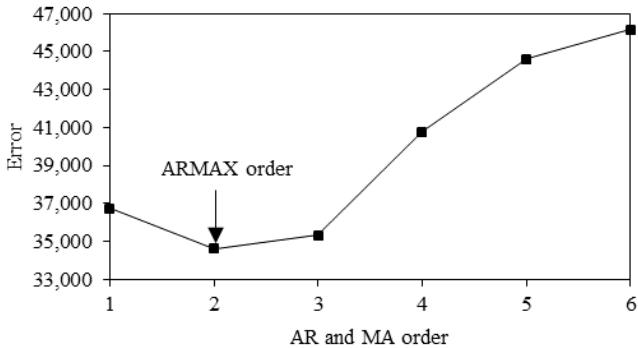

Fig. 5 Determining AR and MA orders based on error value.

### D. Data Post-processing

The resultant forecasted data from last stage represent the daytime GHI values in normalized form. Last step. i.e., data post-processing, includes three steps: denormalization, adding nighttime hours, and calculating the forecasted GHI.

The performance of the model is accordingly evaluated by calculating the Normalized Root Mean Square Error (NRMSE) as in (5).

$$NRMSE = \frac{\sqrt{\frac{1}{N}*\sum_{t=1}^{N}(GHI(t)_{actual}-GHI(t)_{forecast})^2}}{\frac{1}{N}*\sum_{t=1}^{N}GHI(t)_{actual}} \quad (5)$$

### III. NUMERICAL SIMULATIONS

The hourly GHI data for the Denver International Airport are used for forecasting [31]. The proposed model is applied to three test days under different weather conditions, and the $R^2$ and the NRMSE are computed to evaluate output result in stage 1 and 2, respectively. The coefficient $R^2$ is used to evaluate the performance of the fitting model resulted from the NARNN, and accordingly the NRMSE is applied in the ARMAX model to evaluate the efficiency of the forested GHI. In order to present the effectiveness of the proposed two-stage forecasting model and show the role of the data detrending, the following cases are studied:
- Case 1: Forecasting using the proposed model with stationary data.
- Case 2: Forecasting using the proposed model with nonstationary data
- Case 3: Forecasting using only one stage instead of the proposed two-stage method

*Case1: Using Two-Stage Model with Stationary Data.*

In this case, stationary data with the proposed two-stage forecasting model are used for solar forecasting. In this respect, first, the historical GHI data undergo mentioned processes to ensure stationarity before the data are fed to NARNN. Second, the NARNN is trained based on the stationary data to establish the target fitting model. Fig. 6 depicts the fitting model predicted by the NARNN model for a cloudy day. The calculated $R^2$ is 0.90, which reveals that the predicted fitting model is close to the target and the fitting model is well predicted in order to be fed to Stage 2. Third step is to introduce the fitting model to the Stage 2 forecasting, i.e., the ARMAX model. The ARMAX model is developed using order 2 for both AR and MA. It should be noted that the previous day is used to train the model and identify ARMAX coefficient. The ARMAX model forecasts the output as shown in Fig. 7. The NRMSE is calculated as 0.085.

In order to evaluate the efficiency of the proposed model, Case 1 is further applied to two additional days. The predicted fitting model by Stage 1 and the forecasted GHI resulted from Stages 2 for the partly cloudy day are depicted in Figs. 8 and 9, respectively. Similarly, the fitting model and the forecasted GHI for the sunny day are respectively shown in Figs. 10 and 11. Table I summarizes the obtained $R^2$ and the NRMSE for each of the studied days. As tabulated in Table I, the NARNN detects the fitting model and the ARMAX forecasts the GHI quite accurately. Moreover, the proposed two-stage model with stationary data can accurately forecast not only the sunny days, but also the cloudy and partly cloudy ones.

TABLE I
$R^2$ AND NRMSE FOR THE FITTING MODEL AND FORECASTED GHI IN CASE 1.

| Weather Condition | $R^2$ | NRMSE |
|---|---|---|
| Cloudy | 0.90 | 0.085 |
| Partly Cloudy | 0.91 | 0.100 |
| Sunny | 0.86 | 0.048 |

*Case 2: Using Two-Stage Model with Nonstationary Data.*

Data stationarity has a significant role in forecasting solar irradiance. This case aims at investigating the effect of using stationary data in the proposed method. In this regards, the proposed two-stage method is utilized but instead of feeding stationary data to the model, the nonstationary data are used. The simulation processes including Stages 1 and 2 (fitting model predicted by NARNN and the ARMAX model) are completely executed for Case 2 without the pre-processing. Table II compares the NRMSE values in this case with the same three days as in Case 1. The NRMSE values for the two-stage method using nonstationary solar data are higher compared to NRMSE values computed before. That means even though the cascaded two-stage method is a useful approach to deal with nonlinear and linear parts of the forecasting, the data stationarity plays a critical role in the accuracy of the results.

*Case 3: Using One-Stage Forecasting Method.*

To show the merits of the two-stage method over a single-stage method, the forecasting is performed using only one stage, here the NARNN. In this case, the stationary data are applied to forecast the target days. That is, similar to Case 1, the historical GHI data undergo preprocess to ensure stationarity before the data is used in the NARNN. It is then trained based on the historical data set considering a similar number of previous hourly samples, as in Cases 1 and 2. Table II shows the NRMSE values for the single-stage method comparing with the proposed two-stage method in Case 1. The two-stage method in Case 1 has a better performance in solar forecasting as the NRMSE values are much less than this case, exhibiting the advantages of the two-stage method over a single-stage method. These results advocate on the merits of decomposing model to reap the benefit of both linear and nonlinear parts in the proposed model.

TABLE II
THE NRMSE FOR DIFFERENT CASE STUDIES

| Weather Condition | NRMSE (Two-Stage model and stationary data) | NRMSE (Two-Stage model and nonstationary data) | NRMSE (One stage model NARNN) |
|---|---|---|---|
| Cloudy | 0.085 | 0.3007 | 0.512 |
| Partly Cloudy | 0.100 | 0.6799 | 0.9899 |
| Sunny | 0.048 | 0.212 | 0.301 |

The maximum NRMSE for one day ahead forecast using the proposed two-stage model and stationary data under different weather conditions is 0.1, which is a promising result. In the proposed model, the minimum NRMSE is achieved in a sunny day, which is quite expected as the trends of the solar time series under clear sky conditions are more predictable. Nevertheless, the two-stage model accuracy has improved by 71% to 85% when using stationary data compared to nonstationary data. Finally, the two-stage model outperforms the single-stage model and reduces the NRMSE by almost 83% to 90%.

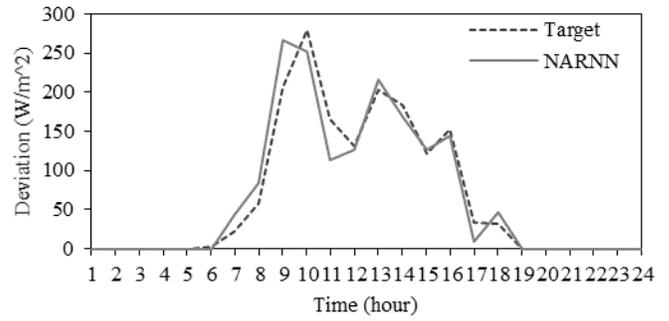
Fig. 6 The fitting model for the cloudy day using NARNN.

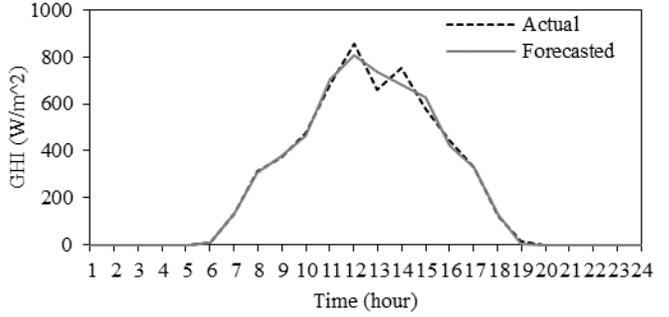
Fig. 7 Forecasted GHI for the cloudy day using ARMAX.

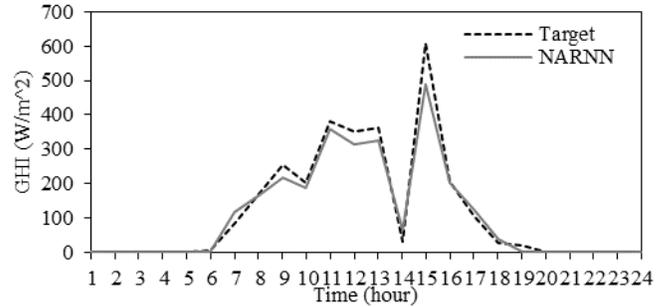
Fig. 8 The fitting model for the partly cloudy day using NARNN.

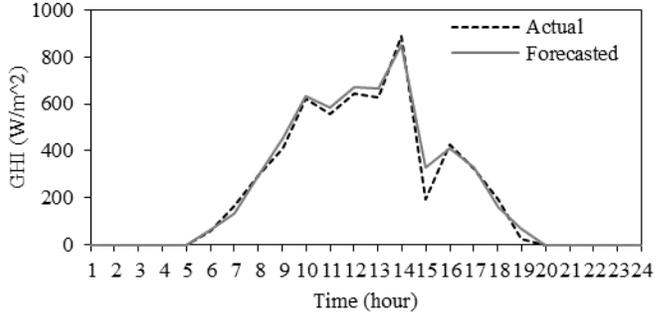
Fig. 9 Forecasted GHI for the partly cloudy using ARMAX.

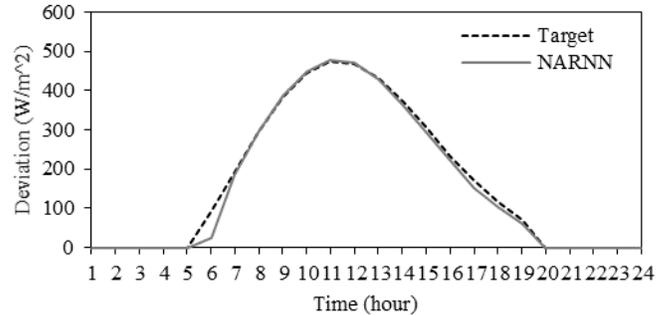
Fig. 10 The fitting model for the sunny day using NARNN.

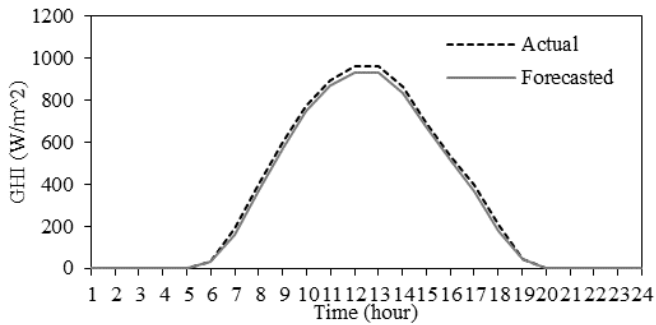

Fig. 11 Forecasted GHI for the sunny day using ARMAX.

## IV. CONCLUSION

In this paper, a solar forecasting model based on decomposed linear and nonlinear statistical methods was proposed. The proposed model benefited from NARNN in Stage 1 forecasting and ARMAX in Stage 2 forecasting combined with a carefully developed data processing approach. The model was simulated to forecast three days under different weather conditions of sunny, partly cloudy, and cloudy. The maximum resultant NRMSE was obtained as 0.1, for a partly cloudy day, which shows the acceptable performance of the proposed model. To exhibit the effectiveness of the two-stage model, three cases were further studied, comparing the two-stage model with a single-stage model, which clearly demonstrated improvements in NRMSE. The importance of the data stationarity in improving forecasting accuracy was moreover investigated.

## REFERENCES

[1] S. Watetakarn and S. Premrudeepreechacharn, "Forecasting of solar irradiance for solar power plants by artificial neural network," in *Smart Grid Technologies-Asia (ISGT ASIA), 2015 IEEE Innovative*, 2015, pp. 1–5.
[2] S. Kann, J. Baca, M. Shiao, C. Honeyman, A. Perea, and S. Rumery, "US Solar Market Insight - Q3 2016 - Executive Summary," GTM Research,Wood Mackenzie Business and the Solar Energy Industries Association.
[3] "IMPACTS OF SOLAR INVESTMENT TAX CREDIT EXTENSION." SEIA, 18-Dec-2015.
[4] A. Tuohy *et al.*, "Solar Forecasting: Methods, Challenges, and Performance," *IEEE Power Energy Mag.*, vol. 13, no. 6, pp. 50–59, Nov. 2015.
[5] J. Kleissl, *Solar Energy Forecasting and Resource Assessment*. Academic Press, 2013.
[6] J. Remund, R. Perez, and E. Lorenz, "Comparison of solar radiation forecasts for the USA," in *Proc. of the 23rd European PV Conference*, 2008, pp. 1–9.
[7] M. Diagne, M. David, P. Lauret, J. Boland, and N. Schmutz, "Review of solar irradiance forecasting methods and a proposition for small-scale insular grids," *Renew. Sustain. Energy Rev.*, vol. 27, pp. 65–76, Nov. 2013.
[8] A. M. Foley, P. G. Leahy, A. Marvuglia, and E. J. McKeogh, "Current methods and advances in forecasting of wind power generation," *Renew. Energy*, vol. 37, no. 1, pp. 1–8, Jan. 2012.
[9] R. Huang, T. Huang, R. Gadh, and N. Li, "Solar Generation Prediction using the ARMA Model in a Laboratory-level Micro-grid," in *Smart Grid Communications (SmartGridComm), 2012 IEEE Third International Conference on*, 2012, pp. 528–533.
[10] G. P. Nason, "Stationary and non-stationary times series," *Stat. Volcanol. Spec. Publ. IAVCEI*, vol. 1, pp. 0–0, 2006.
[11] A. Mellit and S. A. Kalogirou, "Artificial intelligence techniques for photovoltaic applications: A review," *Prog. Energy Combust. Sci.*, vol. 34, no. 5, pp. 574–632, Oct. 2008.
[12] J. M. Filipe, R. J. Bessa, J. Sumaili, R. Tome, and J. N. Sousa, "A hybrid short-term solar power forecasting tool," in *Intelligent System Application to Power Systems (ISAP), 2015 18th International Conference on*, 2015, pp. 1–6.
[13] M. Tucci, E. Crisostomi, G. Giunta, and M. Raugi, "A Multi-Objective Method for Short-Term Load Forecasting in European Countries," *IEEE Trans. Power Syst.*, vol. 31, no. 5, pp. 3537–3547, Sep. 2016.
[14] A. Mellit, M. Benghanem, and S. A. Kalogirou, "An adaptive wavelet-network model for forecasting daily total solar-radiation," *Appl. Energy*, vol. 83, no. 7, pp. 705–722, Jul. 2006.
[15] H. T. C. Pedro and C. F. M. Coimbra, "Assessment of forecasting techniques for solar power production with no exogenous inputs," *Sol. Energy*, vol. 86, no. 7, pp. 2017–2028, Jul. 2012.
[16] C. Paoli, C. Voyant, M. Muselli, and M.-L. Nivet, "Forecasting of preprocessed daily solar radiation time series using neural networks," *Sol. Energy*, vol. 84, no. 12, pp. 2146–2160, Dec. 2010.
[17] D. Yang, P. Jirutitijaroen, and W. M. Walsh, "Hourly solar irradiance time series forecasting using cloud cover index," *Sol. Energy*, vol. 86, no. 12, pp. 3531–3543, Dec. 2012.
[18] R. Marquez and C. F. M. Coimbra, "Intra-hour DNI forecasting based on cloud tracking image analysis," *Sol. Energy*, vol. 91, pp. 327–336, May 2013.
[19] C. W. Chow *et al.*, "Intra-hour forecasting with a total sky imager at the UC San Diego solar energy testbed," *Sol. Energy*, vol. 85, no. 11, pp. 2881–2893, Nov. 2011.
[20] E. Lorenz, J. Hurka, D. Heinemann, and H. G. Beyer, "Irradiance Forecasting for the Power Prediction of Grid-Connected Photovoltaic Systems," *IEEE J. Sel. Top. Appl. Earth Obs. Remote Sens.*, vol. 2, no. 1, pp. 2–10, Mar. 2009.
[21] C. Chen, S. Duan, T. Cai, and B. Liu, "Online 24-h solar power forecasting based on weather type classification using artificial neural network," *Sol. Energy*, vol. 85, no. 11, pp. 2856–2870, Nov. 2011.
[22] P. Bacher, H. Madsen, and H. A. Nielsen, "Online short-term solar power forecasting," *Sol. Energy*, vol. 83, no. 10, pp. 1772–1783, Oct. 2009.
[23] S. Chai, Z. Xu, and W. K. Wong, "Optimal Granule-Based PIs Construction for Solar Irradiance Forecast," *IEEE Trans. Power Syst.*, vol. 31, no. 4, pp. 3332–3333, Jul. 2016.
[24] R. H. Inman, H. T. C. Pedro, and C. F. M. Coimbra, "Solar forecasting methods for renewable energy integration," *Prog. Energy Combust. Sci.*, vol. 39, no. 6, pp. 535–576, Dec. 2013.
[25] H. S. Jang, K. Y. Bae, H.-S. Park, and D. K. Sung, "Solar Power Prediction Based on Satellite Images and Support Vector Machine," *IEEE Trans. Sustain. Energy*, vol. 7, no. 3, pp. 1255–1263, Jul. 2016.
[26] W. Ji and K. C. Chee, "Prediction of hourly solar radiation using a novel hybrid model of ARMA and TDNN," *Sol. Energy*, vol. 85, no. 5, pp. 808–817, May 2011.
[27] K. Benmouiza and A. Cheknane, "Small-scale solar radiation forecasting using ARMA and nonlinear autoregressive neural network models," *Theor. Appl. Climatol.*, vol. 124, no. 3–4, pp. 945–958, May 2016.
[28] M. Alanazi and A. Khodaei, "Day-ahead Solar Forecasting Using Time Series Stationarization and Feed-Forward Neural Network," presented at the North American Power Symposium (NAPS), 2016, Denver, 2016, pp. 1–6.
[29] G. E. P. Box, G. M. Jenkins, and G. C. Reinsel, *Time series analysis: forecasting and control*. Hoboken, N.J.: John Wiley, 2008.
[30] H. Akaike, "A new look at the statistical model identification," *IEEE Trans. Autom. Control*, vol. 19, no. 6, pp. 716–723, 1974.
[31] "Solar Data 1991-2010 Site #725650." [Online]. Available: http://rredc.nrel.gov/solar/old_data/nsrdb/1991-2010/hourly/siteonthefly.cgi?id=725650. [Accessed: 16-Sep-2014].